\documentclass[letterpaper, 10 pt, conference]{ieeeconf}  % Comment this line out if you need a4paper
\IEEEoverridecommandlockouts                              % This command is only needed if 
\usepackage{graphics}
\usepackage{amsmath}
\usepackage{amssymb}
\usepackage{xcolor}
\usepackage{bbold}
\usepackage{bm}
\usepackage{optidef}
\usepackage{mathrsfs}  % script font
\usepackage[noadjust]{cite}
\usepackage{lipsum}
\usepackage{todonotes}
\usepackage{xfrac}  % for \sfrac
\usepackage{bigdelim}
\usepackage{tikz}
\usetikzlibrary{backgrounds, chains}
\usetikzlibrary{arrows.meta}
\usepackage{booktabs}
\usepackage{scalerel}

\definecolor{patrick_color}{rgb}{.0,.6,.05}
\definecolor{gerry_color}{rgb}{.5,.7,.1}
\definecolor{frank_color}{rgb}{0.75,0.25,0.0}
\definecolor{seth_color}{rgb}{0.6,0.0,0.6}

\newcommand{\norm}[1]{\left\lVert#1\right\rVert}
\newcommand{\ff}{{f\kern-0.15em f}} % feedforward

\usepackage{xspace}

\newcommand*{\eg}{e.g.\@\xspace}
\newcommand*{\ie}{i.e.\@\xspace}

\makeatletter
\newcommand*{\etc}{%
    \@ifnextchar{.}%
        {etc}%
        {etc.\@\xspace}%
}
\makeatother

\title{\LARGE \bf
Architectural-Scale Artistic Brush Painting with a Hybrid Cable Robot
}

\author{Gerry Chen$^1$, Tristan Al-Haddad$^2$, Frank Dellaert$^1$, and Seth Hutchinson$^1$\vspace*{-0.15em}%
\thanks{This work was supported by NSF Grant No. 2008302, the Georgia Tech Library's Artist-In-Residence program, and Formations Studio.
}%
\thanks{
    $^1$Institute for Robotics and Intelligent Machines (IRIM), 
Georgia Institute of Technology, Atlanta, GA, USA,
        \scalebox{.9}[1.0]{\tt \{gchen,fd27,seth\}@gatech.edu}
        }%
\thanks{
    $^2$Formations Studio, Atlanta, GA, USA,
        \scalebox{.9}[1.0]{\tt tristan@formations.works}}
}

\makeatletter
\let\NAT@parse\undefined
\makeatother
\usepackage{hyperref}
\hypersetup{
    pdftitle={Architetural-Scale Artistic Painting with a Hybrid Cable-Driven Robot},
    pdfauthor={Gerry Chen and Frank Dellaert and Seth Hutchinson},
    pdfsubject={Robotic Trajectory Retiming},
    pdfkeywords={trajectory retiming, time-optimal path parameterization, TOPP, TOPP-RA, factor graph, variable elimination},
    bookmarksnumbered=true,     
    bookmarksopen=true,         
    bookmarksopenlevel=1,       
    colorlinks=true,
    allcolors=blue,
    pdfstartview=Fit,           
    pdfpagemode=UseOutlines,
    }
\usepackage{hypcap} % link figures to image not caption
\begin{document}

\maketitle
\thispagestyle{empty}
\pagestyle{empty}

%%%%%%%%%%%%%%%%%%%%%%%%%%%%%%%%%%%%%%%%%%%%%%%%%%%%%%%%%%%%%%%%%%%%%%%%%%%%%%%%
\begin{abstract}
Robot art presents an opportunity to both showcase and advance state-of-the-art robotics through the challenging task of creating art.
% Creating large-scale artworks in particular engage the public in a way that small-scale works cannot, but the scale of the artwork also presents unique challenges to robotic systems.
% Inspired by both large-scale mural painting robots and small-scale brush painting robots, our work paints indoors on glass using a brush which requires large-scale yet precise, dextrous motion control control of the brush.
Creating large-scale artworks in particular engages the public in a way that small-scale works cannot, and the distinct qualities of brush strokes contribute to an organic and human-like quality.
Combining the large scale of murals with the strokes of the brush medium presents an especially impactful result, but also introduces unique challenges in maintaining precise, dextrous motion control of the brush across such a large workspace.
In this work, we present the first robot to our knowledge that can paint architectural-scale murals with a brush.  We create a hybrid robot consisting of a cable-driven parallel robot and 4 degree of freedom (DoF) serial manipulator to paint a 27m by 3.7m mural on windows spanning 2-stories of a building.
% Pane width: 72" (mini pane), or 84" (full pane)
% Inter-pane gap: 3"
% 6 full pane pairs, plus 1 half-mini-pane and 1 half-full-pane
%   = (72 + 84 + 6) * 6.5 = 1053" = ~26.75m
% Pane height: 71.5", inter-pane gap: 2.5" -> total height is 145.5" = 3.7m
We discuss our approach to achieving both the scale and accuracy required for brush-painting a mural through a combination of novel mechanical design elements, coordinated planning and control, and on-site calibration algorithms with experimental validations.
\end{abstract}

%%%%%%%%%%%%%%%%%%%%%%%%%%%%%%%%%%%%%%%%%%%%%%%%%%%%%%%%%%%%%%%%%%%%%%%%%%%%%%%%
\section{Introduction} \label{sec:intro}
% For roboticists, it can be easy to overlook the importance of art in inspiring and communicating the future of robotics in society.
Art, including visual arts, music, and dance, is a special medium of communication that can engage and inspire the public in a way that traditional robotics research often struggles to do, while the challenge of robot art simultaneously pushes forward the state-of-the-art in robotics.
For example, the Boston Dynamics robot dance \href{https://youtu.be/fn3KWM1kuAw}{video}
%  of their robots dancing to the song ``Do you love me?''
reached 30 million viewers -- likely more than any academic paper ever written.
For this work, we seek to create a large-scale mural which draws-in audience members to witness live robot painting and inspire learning about robotics for years to come.
% Perhaps the most well-known example of robot art is Boston Dynamics' video\footnote{\url{https://youtu.be/fn3KWM1kuAw}} of their robots dancing to the song ``Do you love me?'', in which a 3 minute video showcases the capabilities of current state-of-the-art robots and simultaneously inspires a view of robot companions that move like friends and pets, all in a uniquely entertaining format that reached 30 million viewers - surely far more than even the most highly cited robotics papers and textbooks.

A number of prior works have explored robotic painting with brushes.  The most common approach is to use painterly rendering algorithms such as \cite{Hertzmann98cgit_npr,Hertzmann03cga_survey-stroke} to generate brush stroke placements which are then executed by a serial manipulator \cite{Schaldenbrand23icra_FRIDA}.  Some more recent approaches build on the traditional painterly rendering approach with learned brush models \cite{Schaldenbrand23icra_FRIDA}, optimization \cite{Wang20iros_RobotCalligraphy}, differentiable rendering \cite{Frans22arxiv_CLIPdraw,Zou20arxiv_StylizedNeuralPainting}, and semantic objective functions \cite{Schaldenbrand22arxiv_StyleCLIPDraw,Vinker22siggraph_CLIPasso} to generate higher quality renderings.  
Other works apply generative models to generate brush strokes directly \cite{Ha18iclr_SketchRNN,Wang23cvm_strokeGANpainter}.
Robot-centric approaches close the loop between the physical artwork and rendering to produce better physical paintings
% More collaborative approaches are notable in closing the loop between the physical artwork and painterly rendering to produce better results when executed on actual robots
\cite{Schaldenbrand23icra_FRIDA,Schaldenbrand24icra_CoFRIDA}.
Nevertheless, both stroke generation algorithms and robot platforms are not conducive to large-scale paintings: the number of strokes generally increases proportionally to the canvas area and serial manipulators are difficult to scale to architectural sizes.

\begin{figure}
  \centering
  \includegraphics[width=\linewidth]{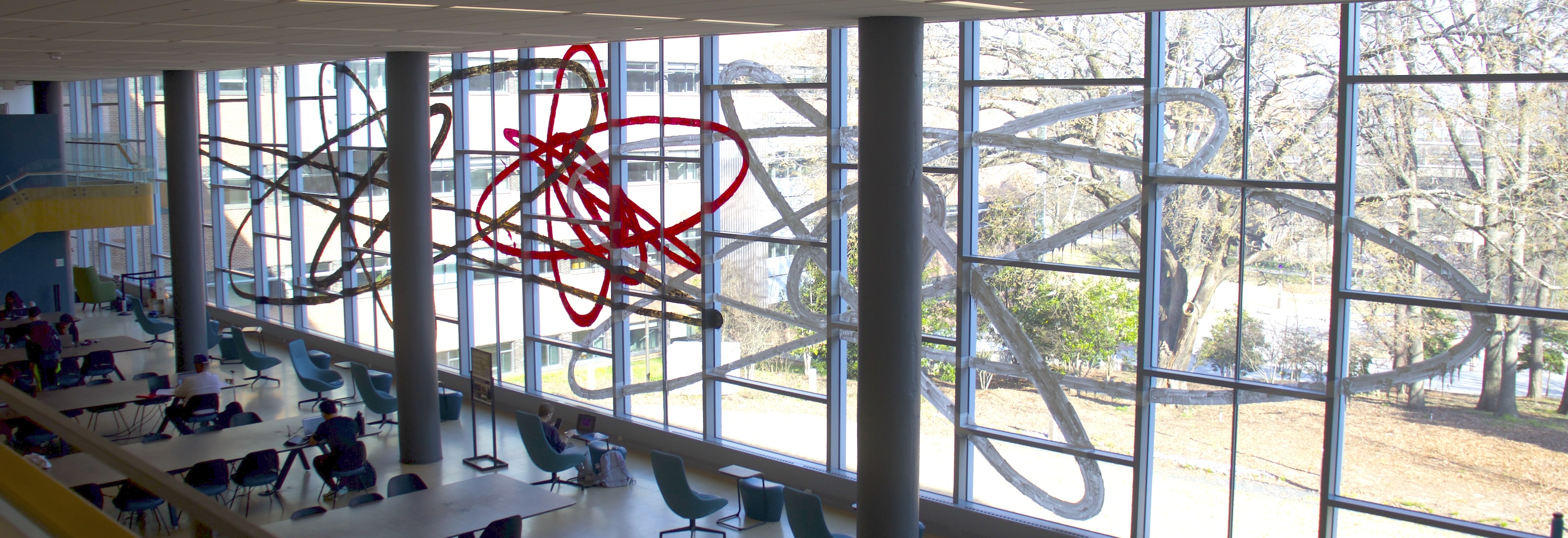}
  \\[0.02\linewidth]
  \includegraphics[width=0.15\linewidth]{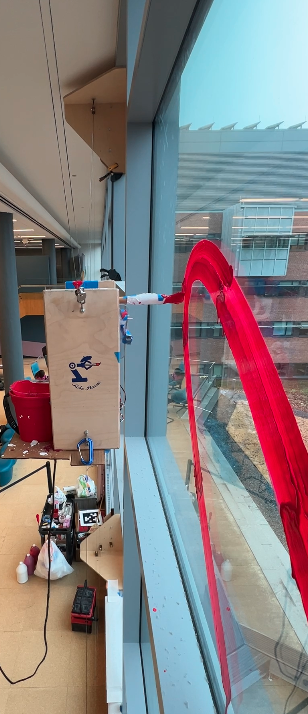}
  % ~\\~
  % ~\\~
  % \includegraphics[width=0.83\linewidth]{figs/headline.jpg}
  \hfill
  \includegraphics[width=0.83\linewidth]{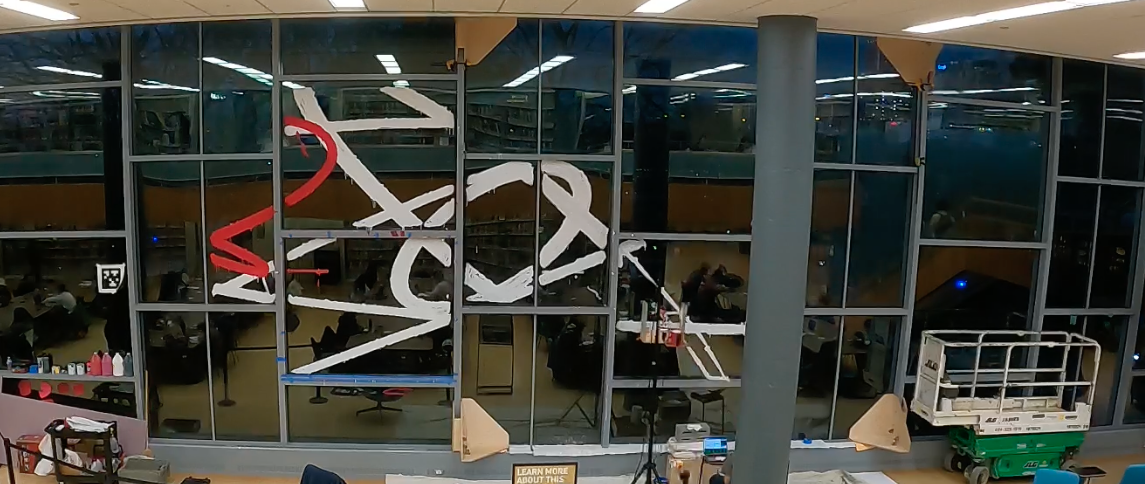}
  \caption{\emph{Polycentric Truthes} [sic] is a 27m by 3.7m mural painted on the windows of the Price Gilbert Library in Atlanta for the Library's Artist-in-Residence program with an audience of thousands of visitors.  The mural was designed by Tristan Al-Haddad and painted by our hybrid cable-driven robot over the course of 7 days across 3 weeks.}
  \label{fig:headline}
\end{figure}

Meanwhile, architectural-scale murals have been painted by robots using other mediums.
Graffiti spray-painted murals have been painted using cable robots \cite{Chen22icra_GTGraffitiSprayPainting,Chen22iros_LocallyOptimalEstimation,Lehni08perspecta_graffitibot_hektor}, UAVs \cite{Uryasheva19siggraph_multiDroneGraffiti,Ratti19web_ufo}, and gantry robots \cite{Roy21youtube_train_writing_graffiti}, but spray paint is not well-suited for indoor environments and existing works have limited capacities for varied stroke shapes, especially compared to brush paintings.  Marker-based graffiti murals have been painted using (humanoid) mobile robots \cite{Jun16iros_Humanoid_Graffiti} and gantry robots \cite{Stuff21youtube_stuff-made-here-painting}, but markers too lack the stroke variation of brushes which give an organic element to the artwork.

In this work, we address the many challenges associated with creating an architectural-scale, brush-painting mural robot.  Specifically, the flexibility of the brush tip requires considerably more precision (especially in the out-of-plane direction) and dexterity than existing large-scale graffiti painting robots, while the large size of the mural makes existing brush painting robot systems impractical.
While many prior works have explored robotic painting of large-scale murals and brush paintings independently, to our knowledge, our work is the first to combine the two.  Achieving this combination of scale and precision requires novel insights into mechanical design, control, and calibration.

% Our contributions include:
% \begin{itemize}
%   \item Analysis of cable routing solutions,
%   \item Design of a brush stabilizing mechanism
%   \item
% \end{itemize}
Our novel contributions include an analysis of CDPR cable routing solutions, the design of a brush stabilizing mechanism, and a joint- and task- space calibration algorithm for registering to absolute on-site canvas locations.
We also demonstrate our robot painting a large-scale mural and share a variety of practical information for robot mural painting.
% A number of practical learnings regarding additional design decisions such as CDPR control, trajectory representations, state management, and safety features are also discussed.

% Finally, we demonstrate our robot's capabilities by painting a 27m by 3.7m mural on windows spanning 2-stories of a building at Georgia Tech.  The mural, titled
% ``Polycentric Truthes'' [sic] was produced as part of the Georgia Tech Library's Artist-in-Residence program to produce a campus artwork installation to engage and inspire passersby.

%%%%%%%%%%%%%
\section*{Approach}
We discuss our system in 3 parts: robot hardware design, software design, calibration/operation.  In \textit{Sec. \ref{sec:approach_hardware}: Robot Hardware Design}, we discuss how the mechanical and electrical design of the robot was chosen to meet the requirements of the painting task.  In \textit{Sec. \ref{sec:approach_software}: Software Design}, we discuss how we control and coordinate the robot to follow a painting trajectory.  In \textit{Sec. \ref{sec:approach_operation}: Calibration, Artwork Design, and Operation}, we discuss how the robot is calibrated, painting trajectories are generated, and the robot is operated.

\section{Robot Hardware Design} \label{sec:approach_hardware}
% * Hardware Design
%   * Arm + CDPR
%     * Large workspace and extensible (can use different paint, different brushes, etc.)
%   * CDPR
%     * overview
%     * doubled-pulleys
%     * anticogging
%   * Arm
%     * overview
%   	* stabilizer horns

% \subsubsection{Arm + CDPR}
% Large workspace and extensible (can use different paint, different brushes, etc.)

Our choice of robot hardware design was motivated by our mural's requirements.  The mural is large, at 27m x 3.7m, planar, and to be painted with a paintbrush in multiple colors.  Furthermore, the mural must be painted without interrupting the library's normal operations, which means that the robot must be able to \textit{safely} paint in a public space.

We chose a hybrid CDPR + serial manipulator design for our painting robot to combine the scalability of CDPRs with the adaptability and dexterity of serial manipulators.
CDPRs, which consist of several winch-actuated cables that pull on a moving platform to control its motion, are ideal for large-scale painting applications since they are characterized by their scalability and high payload capacities.  However, because cables have no ability to resist compression (or bending), the moving platform often suffers from relatively lower stiffness/rigidity. %  - \ie a small disturbance force can cause large deflections.
Fortunately, for the application of painting on a planar surface, contact forces are low and the painting surface can aid in stabilizing the moving platform, together with appropriate internal tensioning of the cables.  Comparing to some alternative large-scale painting platforms studied, cable robots tend to have taller workspaces than mobile robots \cite{Jun16iros_Humanoid_Graffiti}; be cheaper and more portable than gantry systems \cite{Stuff21youtube_stuff-made-here-painting}; and posses greater payloads, easier control, non-battery-limited mission durations, and superior safety than UAVs \cite{Ratti19web_ufo}.
Serial manipulators, on the other hand, are often used in artistic painting robots for their ability to carefully control the brush or marker \cite{Wang20iros_RobotCalligraphy,Schaldenbrand23icra_FRIDA,Berio16iros_dynamic_strokes}.  Although they do not scale to large sizes alone, they have been paired with cable robots in the past to enable dexterity across large workspaces \cite{Chen23icra_hydroponic_robot}.

\begin{figure}
  \centering
  \includegraphics[width=0.6\linewidth,page=9,trim=0 0 7.31in 0,clip]{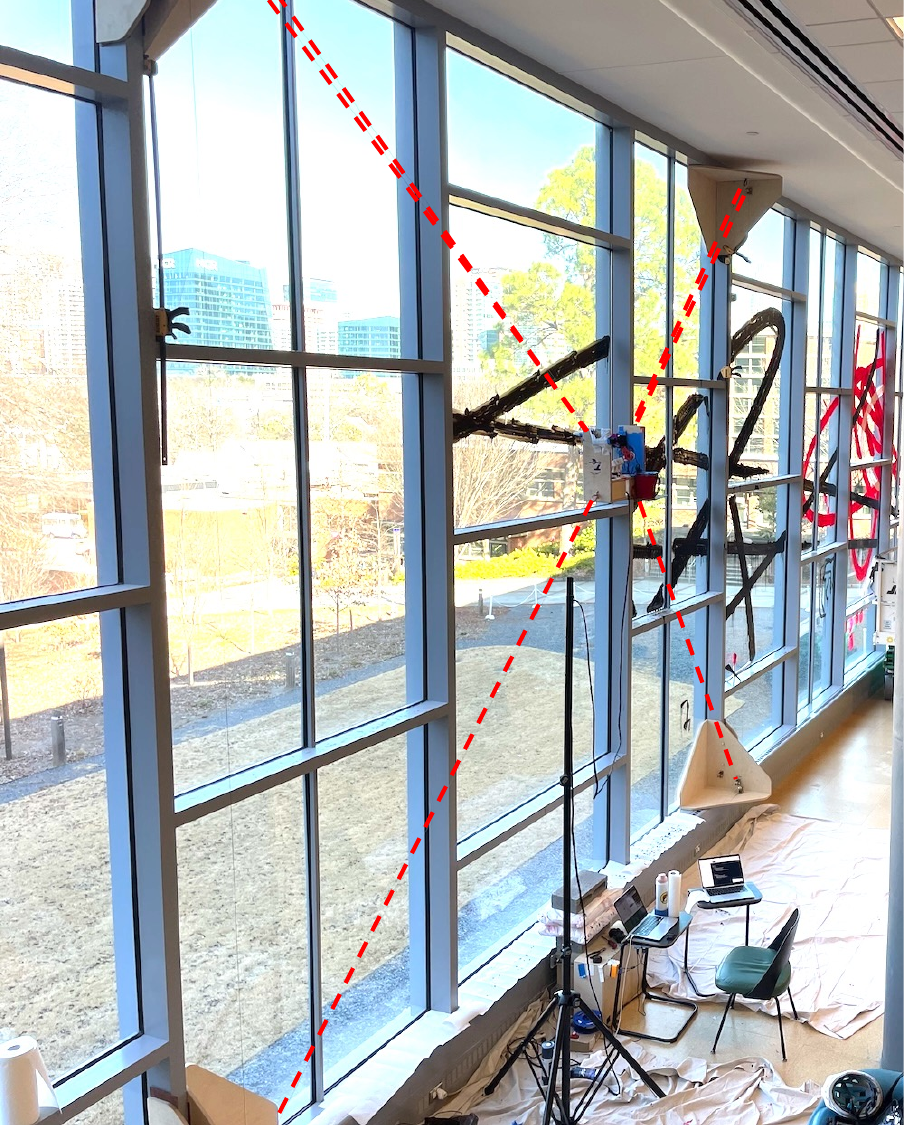}
  \caption{Our 4-cable planar CDPR is 5.8m wide by 3.7m tall.}
  \label{fig:cdpr_overview}
\end{figure}

\begin{figure}
  \centering
  \includegraphics[width=0.5\linewidth,trim={0 4.62in 9.96in 0},clip]{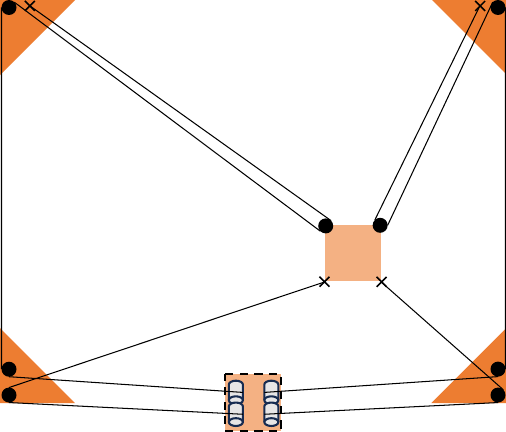}
  \caption{The 4-cable planar CDPR uses platform-mounted pulleys on the top two cables to achieve a 2:1 mechanical force advantage.  We perform an analysis showing we do not lose estimation accuracy with this approach compared to halving the winch radius.}
  \label{fig:cdpr_diagram}
\end{figure}

\subsection{CDPR Design} \label{ssec:cdpr_hardware_design}
\begin{figure}
  \centering
  \includegraphics[width=0.55\linewidth]{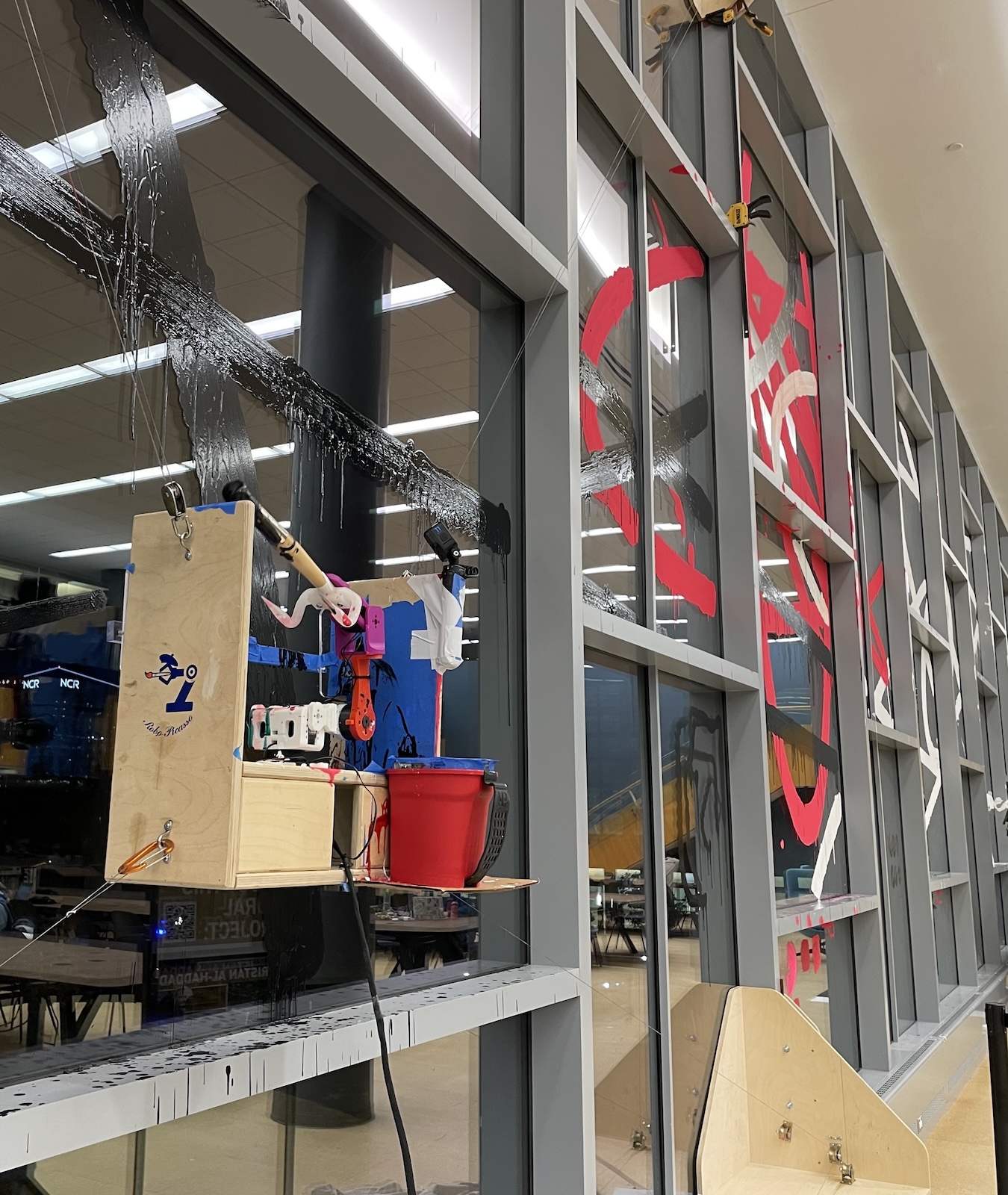}\hfill
  \raisebox{2.7em}{\includegraphics[width=0.38\linewidth,page=8,trim=0 1.88in 8.49in 0,clip]{figs/slides.pdf}}
  \caption{The serial manipulator is mounted on the CDPR's moving platform (Left) controlled by 4 cables routed through pulleys rigidly mounted to the window mullions with wooden clamps (Right).}
  \label{fig:CDPR_mounting_and_arm}
\end{figure}

Our CDPR was designed around the mural's setting, but is flexible and extensible to other settings as well.  We chose a 4-cable planar CDPR 5.8m wide and 3.7m tall (Fig. \ref{fig:cdpr_overview}) and paint in sections, relocating the robot between each section.  4 pulleys are located at the 4 corners of the robot, rigidly mounted to the building's structure with a custom tetrahedral wooden clamp which clamps onto mullions without damaging them (Fig. \ref{fig:CDPR_mounting_and_arm}).  The pulleys are located at 4 corners of a rectangle, and the mounting locations on the moving platform are also at the 4 corners of a rectangle (Fig. \ref{fig:cdpr_diagram}).  Although this results in many singularities which primarily manifest as an inability to control the orientation of the moving platform, we find that the configuration is more passively stable and therefore easier to manage than one which requires feedback on the moving base orientation.

To support the heavy wooden platform with serial manipulator, we seek to achieve more cable tension while minimizing cable length estimation error.  
% As we are concerned about the proprioceptive cable length estimation accuracy, w
We perform a theoretical analysis to compare how the accuracy changes if we add a pulley, half the winch radius, or add a motor gear ratio.
Surprisingly, we have not found such prior analysis in the CDPR literature.
Making the assumption that the majority of the cable length estimation error comes from %variations in cable winding causing 
variations in the effective winch radius, we model how the cable length estimate changes with changes in diameter.  The cable length for the gearbox can be modeled as $\frac{1}{2}\theta d$, so the derivative with respect to winch diameter is $\frac{1}{2}\theta$, where $\theta$ is the output shaft angle and $d$ is the winch diameter.  If we instead half the winch radius, we have the derivative is still $\frac{1}{2}\theta$, but now $\theta$ must be twice as large to elicit the same change robot motion so accuracy halves.  For the pulley, we have the same expression $\frac{1}{2}\theta$ and accuracy halves for the same reason.
We experimentally validate this conclusion in Sec. \ref{ssec:exp_cable_routing}.

For our setup, we opted for a pulley configuration because the winch diameter was already very small at 10mm and motor gearing would increase cost and complexity while also introducing the potential for decreased backdrivability.
Specifically, we employ a 2:1 pulley system for the top two cables of the cable robot.  Instead of mounting a top cable directly to the moving platform, an additional pulley is instead mounted to the moving platform and the cable is routed through this pulley then affixed to the tetrahedral clamp near the fixed-based-pulley as shown in Fig. \ref{fig:cdpr_diagram}.

% In addition to doubling the force applied to the platform, this approach also enables roughly doubling the accuracy of the position estimation as compared to decreasing winch radius.
% This is because we observe that the greatest position inaccuracies arise from estimating cable length given winch rotation angle which we show experimentally becomes more inaccurate with smaller winch radiuses.
% Because we observe that the greatest position inaccuracies arise from estimating cable length from winch rotation angle (which relies on the effective winch radius and therefore the precise winding behavior of the cable), this pulley configuration for mechanical advantage is preferable to either halving the winch radius (which degrades the accuracy by at least 2x) or introducing a motor gear ratio (which is not expected to affect accuracy).

To drive the winches, we use BLDC motors with anti-cogging compensation, which enables smooth, precise control and accurate torque feedback.
% To drive the winches, we use BLDC motors which enable precise control and accurate torque feedback but present challenges in the form of cogging torque.  Cogging torque is a phenomenon in which magnets in the motor's rotor are attracted to the stator's ferrous core in certain positions.  The result is an additional torque that the motor must overcome to move past these positions.  To mitigate this, we apply an anti-cogging algorithm which first models the cogging torque by mapping the torque required to maintain zero velocity at every encoder position, then adding the cogging torque as a feedforward term to torque commands.  Although there are limitations to the approach, chief of which being magnetic field interference among nearby motors and extreme sensitivity to encoder position, we find that this approach is sufficiently effective at reducing the cogging torque to a level that is manageable for our application.

\subsection{Serial Manipulator Design} \label{ssec:hardware_arm}
The serial manipulator is primarily designed around the choice of brush painting as the artistic medium.  The brush is a versatile tool that can be used to create a wide variety of strokes, textures, and effects, while also being a tool that is familiar which we hope will make the mural more relatable to the public.  Finally, as opposed to a more ``industrial'' design (\eg with linear actuators and mechanical linkages), the serial manipulator is designed to be able to manipulate the brush in a way that is similar to how a human artist would manipulate a brush, thereby contributing to the performative aspect of the robot's artistic expression.% of the robot itself.

\begin{figure}
  \centering
  \includegraphics[width=0.5\linewidth,page=2,trim=0 0 8.11in 0,clip]{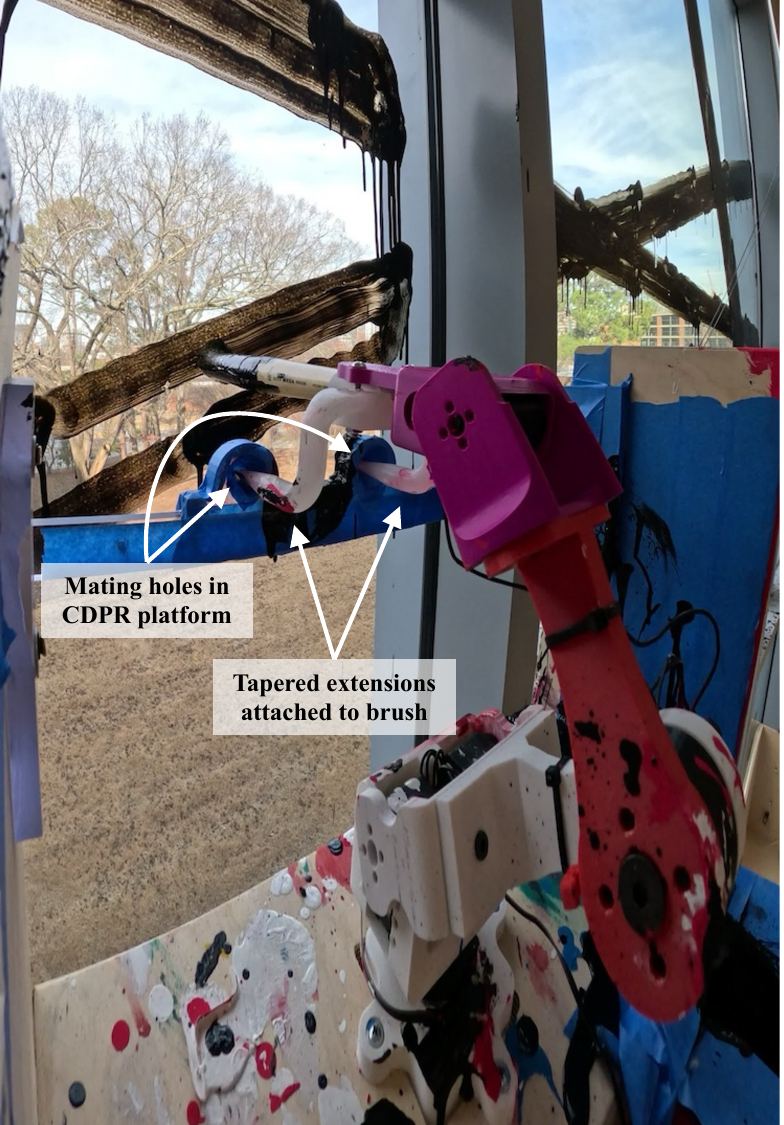}
  \caption{The brush stabilizer's self-centering mates are engaged when the brush is in the painting position.}
  \label{fig:horns}
\end{figure}

To achieve the requisite dexterity to paint, we designed a 4-DoF serial manipulator with a 0.38m reach and measured 150g payload capacity (at the tip) to hold the 110g brush.  The manipulator is designed to be lightweight and affordable, leading to the choice to use AX-12 Dynamixel servos.  Although for this mural we keep the brush perpendicular to the canvas when painting, the 4-DoF serial manipulator complements the CDPR's 2 translational degrees of freedom theoretically enabling SE(3) brush motions.  We use a standard 2-DoF shoulder, 1-DoF elbow, 1-DoF wrist design for the manipulator due to its effectiveness and ubiquity as shown in Fig \ref{fig:horns}.  The second degree of freedom in the shoulder uses two synchronized servos to improve the payload capability of the manipulator.

Due to the significant inaccuracy of the AX-12 servos, we designed a stabilizing stand to keep the manipulator rigid while painting (Fig. \ref{fig:horns}).  During operation, we found that AX-12 servos experience error due to both mechanical backlash and controller degradation after prolonged continuous operation.  As a result, brush positioning suffers from poor accuracy and, more importantly, deflection, which results in ``cutting corners''.  To mitigate this, we designed a stabilizing mechanism whereby a pair of conical extensions on the brush/manipulator end-effector creates tapered fits with circular hole receptacles on the moving platform.  The mechanism is designed such that it is engaged (locks the pose of the brush relative to the platform) when the brush is in the painting position (in contact with the canvas) and can be unengaged by simply pulling the brush back off the canvas with the manipulator.
The mate is designed with a tapered, self-centering fit to be robust to positioning errors as the serial manipulator pushes the brush forward to engage the fit.  The mechanism, engaged, is shown in Fig. \ref{fig:horns}.

Although the stabilizing stand means that the arm cannot move while the painting, we find that this is necessary to attain sufficient accuracy.  We will further discuss in Sec. \ref{sec:approach_software} the tradeoff between producing coordinated motions (\ie CDPR and arm move together while drawing) vs fixing the arm in place while the CDPR moves to paint strokes.

\begin{figure}
  \centering
  \includegraphics[width=0.9\linewidth,page=3,trim=0 3.75in 2.06in 0,clip]{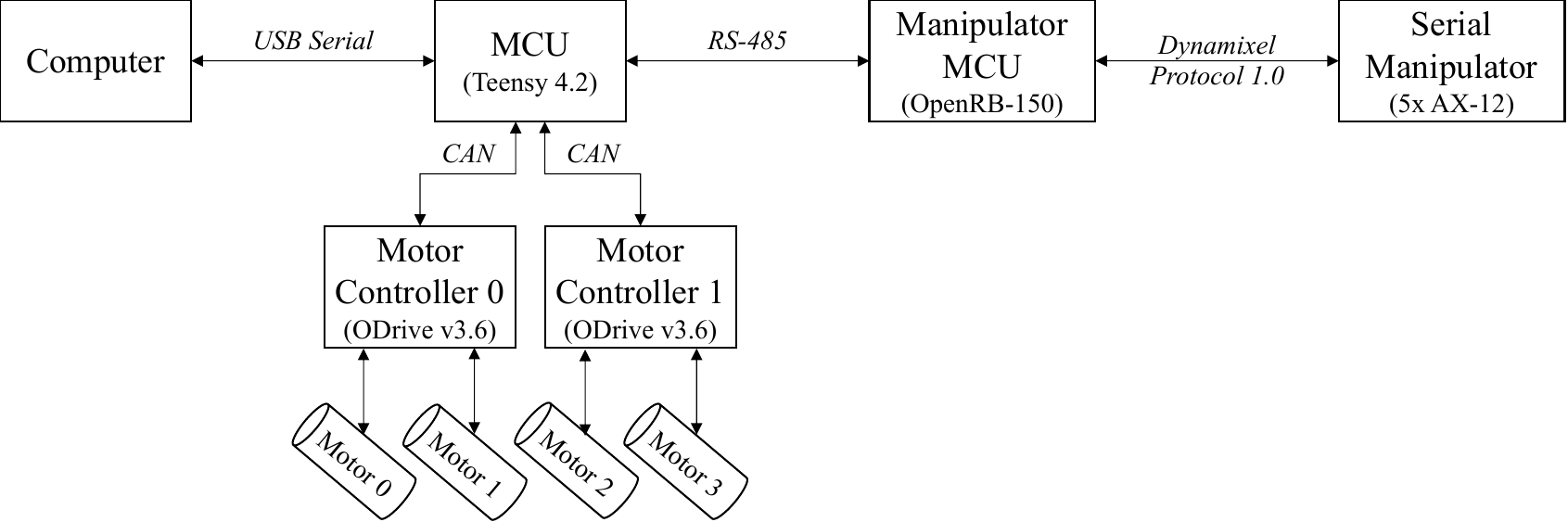}
  \caption{The robot system components communicate with each other over robust interfaces to manage both the CDPR and serial manipulator.}
  \label{fig:electrical_block_diagram}
\end{figure}

\section{Software Design} \label{sec:approach_software}
% * Software Design
%   * CDPR
%     * tracking control
%     * safety stops
%   * Arm
%     * IK & position training
%     * Auto-dipping
%     * Temperature control
%   * Coordination
%     * Pause before painting to arrest motion
%     * Paint switching?
%     * Painting duration estimation
%     * Waypoint vs Line vs spline following code

To coordinate our hybrid robot, we implement low-level controls for the CDPR and arm then tie them together with higher level coordination logic.  For the purposes of this section, it is assumed that a time-stamped painting trajectory is already provided, using the procedures in Sec. \ref{sec:approach_operation}.

% We implement manual and painting control modes, with the manual mode primarily to run calibration and testing, and the painting mode to execute the painting task.  
% \todo{actually, move the initial introduction of ``manual mode'' to the operation section}

\subsubsection*{Decoupled vs coordinated motion}

We largely rely on the CDPR to follow the painting trajectory and reserve the serial manipulator just for moving the brush to make/break contact with the canvas and handle other brush management tasks (\eg dipping to refill paint).
The primary reason is the aforementioned inaccuracy of the hobby-grade serial manipulator which, when not held in place using our stabilizing mechanism, deviates far too much from the desired trajectory to outweigh the benefits of using it for trajectory following.  Furthermore, as the CDPR can smoothly reach the entire canvas and we do not explicitly model how orientation affects brush strokes, there is limited added benefit to moving the manipulator together with the CDPR while painting aside from the performative aspect.
%The reasons are further explained in Sec. \ref{ssec:state_management}.
% The primary purpose of the CDPR is to move the moving platform in a desired painting trajectory while the manipulator is primarily used to move the brush among painting, dipping, and other positions relative to the moving platform.

% \todo{actually, explain the tradeoff between de-coupled arm/cdpr motion vs mobile manipulator math here}

\subsection{CDPR Control} \label{ssec:cdpr_control}

We implement an iLQG tracking controller based on \cite{Chen22iros_LocallyOptimalEstimation}, which receives a desired trajectory and computes tensions to apply to the cables to track the trajectory given cable length and velocity measurements.
The controller consists of an offline stage in which we pre-compute time-varying linear control gains and state estimation gains, and an online stage in which we execute the time-varying, locally-optimal linear state estimator and linear controller.

Offline, our algorithm takes as input a desired painting trajectory and first solves for the optimal nominal trajectory.  The nominal trajectory is computed as the solution to a QP with quadratic state error and control cost objectives, a system model (equality constraints), and control limits (inequality constraints).  The QP has a banded structure and is solved in linear time using iLQG, implemented using GTSAM \cite{Dellaert17fnt}.
Then, the QP problem is linearized around the nominal trajectory and the locally optimal, time-varying linear control gains and Kalman estimation gains are computed.  In GTSAM, the time-varying LQR gains are computed automatically during the last iLQG iteration \cite{Yang21icra_ecLQR} so they can be directly retrieved, while the Kalman estimation gains are computed using a backward-forward algorithm.

The resulting time-varying gains are uploaded to the microcontroller (MCU) then executed in real-time as:
\begin{align}
  \delta \hat{x}_k &= {}^x\! K_k \delta \hat{x}_{k-1} + {}^z\! K_k z_k + k_k \label{eq:estimator} \\
  u_k &= K_k \delta \hat{x}_k + u_k^* \label{eq:controller}
\end{align}
where $\delta \hat{x}_k$ is the estimated deviation of the state from nominal; $z_k$ is the measurement (cable lengths and velocities); $u_k$ is the commanded cable tensions, ${}^x\! K_k$, ${}^z\! K_k$, and $k_k$ are precomputed matrices/vectors for the time-varying linear state estimator, and $K_k$ and $u_k^*$ are precomputed feedback control gains and feedforward controls, respectively.

It may come as a surprise that the optimal state estimator does not appear to follow the standard form of a Kalman filter (while the control gains take the form of a typical time-varying LQR controller).  This is because knowing the nominal trajectory in advance allows us to simplify the online computation by pre-multiplying matrices.
Although full details of the derivation are provided in \cite{Chen22iros_LocallyOptimalEstimation}, the intuition is that the covariances depend only on the system model (which is fixed once we linearize around the nominal trajectory) and not on the real-time measurements so we can pre-compute and combine the predict and update steps into a single, affine function of the previous state estimate and current measurement per timestep.

Given the simplicity of the expressions in Eqs. \eqref{eq:estimator} and \eqref{eq:controller}, the control loop runs at 1 kHz on the MCU.

Occasionally, we must manually control the robot live (\ie with a joystick) such as for setup and testing.  In these cases, the trajectory is not known in advance so the dual-space PID controller from \cite{Gouttefarde15tro} is implemented.  Although the accuracy and dynamic performance of the iLQG controller is superior, this manual mode is used primarily for visual alignment and testing so its reduced performance is acceptable.

\subsection{Serial Manipulator (Arm) Control}

% To satisfy the long-duration painting requirements of the hobby-grade manipulator, we discuss the joint angle training, temperature control, and auto-dipping algorithms.

\begin{figure}
  \centering
  \includegraphics[width=0.65\linewidth,page=4,trim=0 5.49in 8.32in 0,clip]{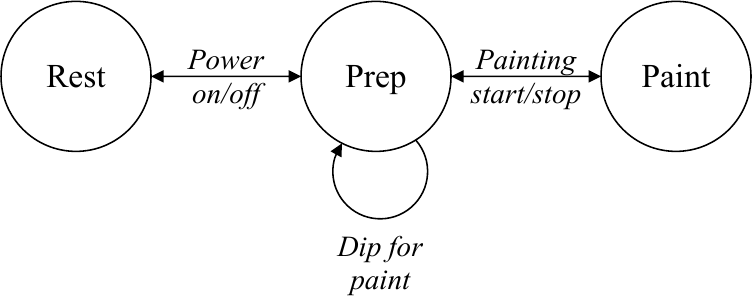}
  \caption{Arm state machine consists of 3 states / configurations: rest (power-off configuration), prep (``central'' configuration not yet touching the canvas but ready to paint), and painting (touching the canvas).}
  \label{fig:arm_state_machine}
\end{figure}

The primary arm functions we seek to implement are:
\begin{itemize}
  \item rest/off
  \item painting (touching the canvas)
  \item refilling (dip the brush into the paint).
\end{itemize}
Accordingly, we can define a set of only 3 configurations and the 3 (bidirectional) trajectories transitioning between them, as shown in Fig. \ref{fig:arm_state_machine}, to cover all the required arm functions.  The ``prep'' state is helpful since represents a configuration where it is relatively easy to find collision-free trajectories to any other configuration the arm needs to reach.  Since the prep state does not make contact with the canvas, we also use it during travel motions.

To program the arm configurations, we take two approaches: manual ``training'' and analytical inverse kinematics.  We find that a simple ``training'' procedure is often easier given that there are only a few configurations required (including collision-free intermediate waypoints) and the servo inaccuracy makes it such that task-space target positions would need to be tuned anyway.  Specifically, the training procedure consists of disabling the servos and manually moving the arm to the desired configuration, then recording the joint angles which are later used as target configurations.  We use this training procedure for the rest configuration and refilling (paint dipping) trajectory.  However, we use the analytical inverse kinematics for the ``prep'' configuration and ``start/stop painting'' trajectories because programming the arm to successfully engage the stabilizing mechanism is easier when adjustments are made in task space (\eg move 2cm higher).  Furthermore, described in Sec. \ref{ssec:hardware_arm}, the servo accuracy degrades with prolonged operation so we compensate by commanding progressively higher (1cm/hr) target ``paint'' positions in task space as the painting progresses.

\begin{figure}
  \centering
  \includegraphics[width=\linewidth,page=5,trim=0 4.21in 2.87in 0,clip]{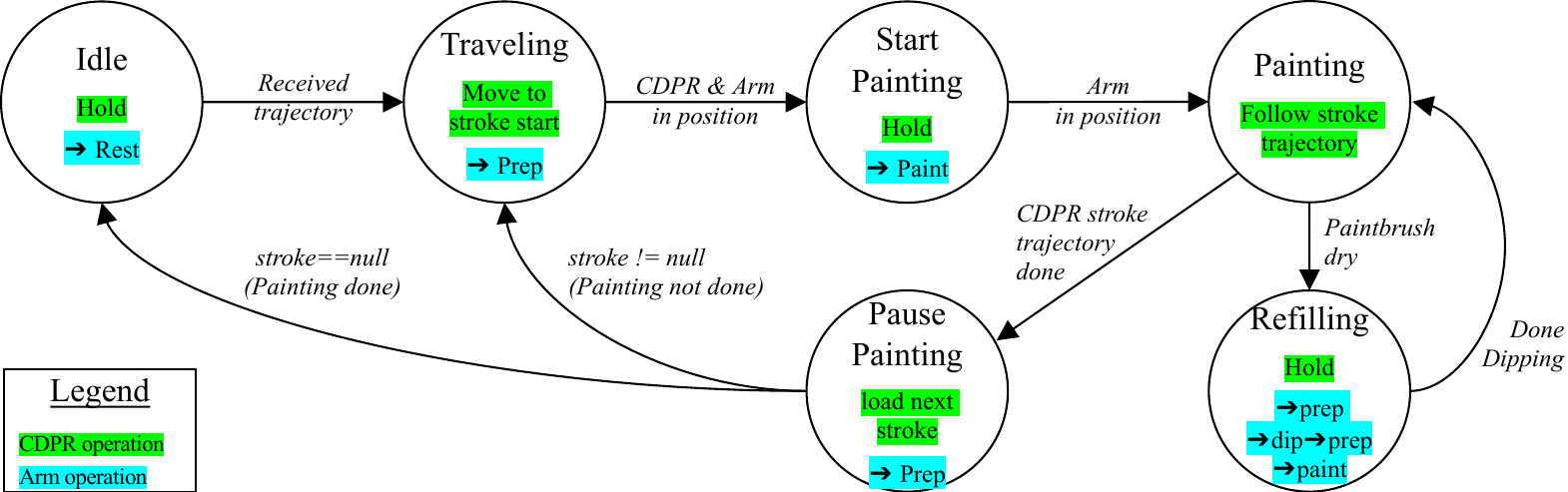}
  \caption{The CDPR + arm pair is coordinated by this state machine to paint.}
  \label{fig:overall_state_machine}
\end{figure}

\subsection{Coordination and State Management} \label{ssec:state_management}

The high-level coordination and state management between the CDPR and Arm is shown in Fig. \ref{fig:overall_state_machine}, and we now also discuss a few non-obvious details.  For the ``\textit{Paintbrush dry}'' condition, we find that distance traveled along the canvas is a good heuristic and we automatically re-dip the brush for every 0.5m the brush has traveled along the canvas.  We also pause slightly (1s and 2s, respectively) just after the transitions from ``Traveling''$\rightarrow$``Start Painting'' and from ``Start Painting''$\rightarrow$``Painting'' to allow the CDPR's motion to arrest, since arm motions often cause slight out-of-plane oscillations in the CDPR platform.  We load each paint color as a separate trajectory so that the robot returns to the ``Idle'' state after each color, enabling us to switch paint colors.

% Although future work could automate the paint switching process, w
For this work, we chose manual paint color switching over automation for three key reasons.
First, cleaning brushes sufficiently to avoid color contamination is impractical -- even humans require 10-15 minutes, so we simply purchase a separate brush for each color.
Second, the drying time of the paint is long and variable, so we intentionally plan the trajectories to minimize color changes.
Finally, the human operators can switch the paint color in less than a minute so the additional complexity of automatic color switching is difficult to justify.
Future works which use different brushes and paints may find automatic paint switching more practical.

\subsection{Safety Controls}

In addition to standard safety measures such as E-Stop buttons, actuator limits, and safety interlocks, we implement a few additional safety features.  

The serial manipulator's hobby-grade servos frequently overheat causing accuracy degradation and, if extreme, automatic shut-down of the servo which may affect the painting or drop objects onto passersby.  In response, we continously monitor servo temperatures and automatically pause painting when the temperature reaches a threshold (65$^\circ$C), wait for the temperature to cool to a safe level, then resume painting.

The CDPR's iLQR controller is also very reliable, but we nonetheless implement safety stops in the event of anomalies to prevent negatively affecting the painting and objects falling onto passersby.
% The primary conditions under which these are activated are when we are in ``manual control mode'', in which case the dual-space PID controller from Sec. \ref{ssec:cdpr_control} is used, when obstacles interfere with the cable routing (e.g. passerby intentionally pulls a cable), or when the base containing the winches is moved from its normal location.
In these cases, a soft limit enacts when the CDPR is 10cm from its commanded position and a hard limit enacts at 20cm.  The soft limit pauses the trajectory but holds the current position while the hard limit immediately stops the trajectory and gently lets the platform down by switching to a gravity compensation mode. % These modes are often preferable to the E-stop button behavior, which shuts off power and causes the platform to freefall due to the lack of a normally-closing brake on the winches.

\section{Calibration, Artwork Design, and Operation} \label{sec:approach_operation}
In this section, we discuss the procedure necessary to operate the robot on-site.  Because the robot is designed to paint 5.8m at a time before needing to be repositioned, calibration is a critical step to ensure that adjacent sections of the mural are painted seamlessly.  We also discuss the painting trajectory generation process and robot operation.

\subsection{Calibration} \label{subsec:calibration}
A number of both proprioceptive and exteroceptive methods for calibration have been studied in the CDPR literature.  Proprioceptive methods leverage the over-actuated quality of CDPRs together with kinematic and dynamic constraints to derive calibration parameters.  Meanwhile, exteroceptive methods leverage external sensors to measure the robot's configuration which often beget more accurate calibration parameters.  In this work, we use a combination of both methods to calibrate the robot, in particular leveraging the architecture of the site itself to provide exteroceptive information.

Our novel approach is to combine proprioceptive calibration for control parameters with exteroceptive calibration for task-space corrections.  Specifically, our proprioceptive calibration involves measuring motor angles during an exploratory motion phase (which is relatively standard in the literature).  Meanwhile, our exteroceptive calibration leverages \textit{operator} feedback to manually position the robot at known locations on the canvas, based on the site's architecture, to provide absolute task space calibration (for which we apply a novel CDPR task-space warping algorithm).

\subsubsection{Proprioceptive calibration} We perform the proprioceptive calibration in two stages.  In the first stage, the operator manually moves the robot to obtain a rough initialization on parameters with which the robot can then use to move itself to a broader set of configurations in the second stage.  This two-stage approach improves safety, since the robot is not required to control itself without a good parameter initialization, while also being quick, easy, and accurate.  Compared to auto-calibration approaches which initialize using small perturbations around the power-on configuration of the robot \cite{Borgstrom09tro_cdpr_self_calibration_displacementjitterdrift}, we find that our approach is more robust to the starting configuration of the robot since it is often powered on very close to the bottom edge of the workspace (due to the large height of the workspace) which results in potentially dangerous conditions for robot actuation.  Other than the procedure used to collect data, the optimization problem used to solve for the calibration parameters is the same for both stages and standard in the literature \cite{Miermeister12gcr_cdpr_autocalib,Chen22iros_LocallyOptimalEstimation}.

During the first proprioceptive calibration stage, the robot is placed in a compliant state by commanding an equal and relatively small tension in all cables allowing the operator to manually move the robot across various configurations reachable by the operator while recording motor angles.  The operator moves the robot to a diverse set of configurations reachable by person for 3 minutes.  The collected data is then used to formulate a nonlinear least squares optimization problem to solve the geometric calibration parameters:
\begin{align}
  \min_{p, \bm{x}} \sum_{k=1}^{n} \norm{l(\theta_k;p)-\hat{l}(A-(x_k+B))}^2 \label{eq:calibration_optimization}
\end{align}
where $l(\theta_k;p)$ computes the cable lengths given measured winch rotations $\theta_k\in\mathbb{R}^4$ at time-step $k$ and parameterized on winch calibration parameters $p$; $\hat{l}(\cdot):=\norm{\cdot}_2$ is the euclidean distance function predicting the cable length; $A\in\mathbb{R}^{4\times2}$ is the locations of the ``base'' pulleys (mounted to the window mullions as in Fig \ref{fig:CDPR_mounting_and_arm}-right) in the world coordinate frame; $B_i\in\mathbb{R}^{4\times2}$ is the location of the platform mounting/pulley location for cable $i$ in the moving platform frame; and $x_k\in\mathbb{R}^2$ is the platform position at time-step $k$ (out of $n$ total time-steps) in the world frame.  As discussed in Sec. \ref{ssec:cdpr_hardware_design}, we chose a configuration with a relatively passively stable platform orientation at the expense of not being able to control or sense orientation well.  Therefore, we assume the orientation to always be the identity rotation so platform pose can be expresed as a position, $x_k\in\mathbb{R}^2$, and can be (broadcast) summed with $B$ for the 4 cables to compute the platform mounting/pulley locations in the world frame.

We choose a quadratic cable length vs winch rotation angle model $l(\theta;p)=p_0+p_1\theta+p_2\theta^2$ with $p\in\mathbb{R}^{3\times 4}$ for the 4 cables to account for the increasing winch radius as the cable winds back on itself on the winch.  Although we have found the true function $l$ to be almost perfectly piecewise linear (each layer of cable winding represents a discrete transition) on data collected with a motion capture system from a smaller-scale test system, we find the quadratic model to be much easier to optimize and still reasonably accurate.

When solving \eqref{eq:calibration_optimization}, we initialize $p$ as $p_2=0$, $p_1$ by the nominal winch radius, and $p_0$ by measuring the initial cable lengths.
Both $A$ and $B$ are assumed to be known and fixed according to the site architecture and platform design.
% since manually measuring them is roughly an order of magnitude more accurate than the error observed in the winch model $l(\theta_k;p)$.

During the second proprioceptive calibration stage, the winch parameters $p$ obtained from the first stage are used to actuate the robot in a grid pattern around the workspace.  The measured motor angles during this motion are then used to solve \eqref{eq:calibration_optimization} again to refine the calibration parameters.  %The resulting parameters are sufficient to robustly control the robot after remounting the robot to each canvas segment.

\subsubsection{Exteroceptive calibration}
We perform the exteroceptive calibration by positioning the robot at known locations on the canvas and computing task-space corrections needed to reach these positions accurately.  This is convenient in many applications since the locations for which accuracy is most critical are often also clearly marked.
For example, in a pick-and-place task, CDPR accuracy is typically most critical at the pick and place locations (which often have clear landmarks) and least critical during the move.
In our application, the window mullions represent the most important locations at which accuracy is required both to prevent damage to the building and to ensure proper alignment across different sections of the mural, which is painting in sections along mullion edges.  Conveniently, the mullions also make for easily accessible locations for the operator to measure.

\begin{figure}
  \centering
  \includegraphics[width=0.5\textwidth,page=7,trim=0 4.08in 4.54in 0,clip]{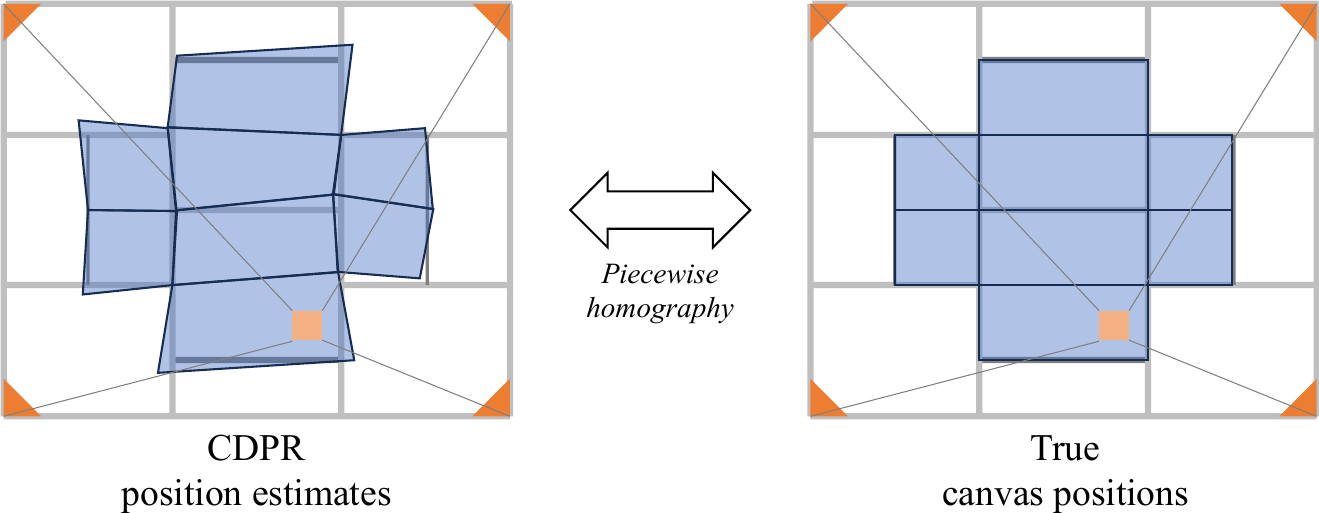}
  \caption{We apply a piecewise homography in the task space to correct the proprioceptively calibrated cable robot's motions to align with the true canvas sections.  Inaccuracy of CDPR estimates is exaggerated for effect.}
  \label{fig:homography_diagram}
\end{figure}

We select 16 mullion intersection locations across the canvas (see Fig. \ref{fig:homography_diagram}).  Next, we have the operator use a joystick to move the robot to each of these locations and record the robot's estimated task-space position.  The true locations are taken from the architectural measurements and verified with a laser distance meter.% which is simplified by the architecturally regular grid with round number dimensions.

To perform calibration, we apply homography transforms\footnote{For details on homographies (projective transforms in computer vision) and how to compute them, see \eg \cite[Ch. 2.1.1 ``Projective'']{Szeliski22book_ComputerVision}.} in the task space to each of the rectangular segments of the workspace defined by the calibrated positions.  With the 16 points measured in our application, we form 8 rectangular sections of the painting portion of the canvas (Fig. \ref{fig:homography_diagram}).  For each section, we compute a $3\times 3$ homography matrix, $H$, mapping the true position to the estimated cable robot position using the 4 corner correspondences:
\begin{align}
  \begin{bmatrix}x_{cdpr}\\y_{cdpr}\\1\end{bmatrix} &= \frac{1}{z}H \begin{bmatrix}x_{true}\\y_{true}\\1\end{bmatrix} \label{eq:homography}
\end{align}
where $(x_{cdpr}, y_{cdpr})$ and $(x_{true}, y_{true})$ are the cable robot's estimated and the true canvas position, respectively; and $z$ is a scaling factor to normalize the homogeneous vectors.

When painting, we simply apply the appropriate homography (based on which rectangular section we are in) to all the commanded trajectories to obtain corresponding trajectories in the CDPR's ``erroneous'' coordinate frame which is used in closed-loop control.  Because adjacent rectangular sections share edges, transitions between homographies are guaranteed to be continuous.  As compared to applying the inverse homography to the CDPR's internal forward kinematics state estimation inside the control loop, operating the control loop on the non-homography-corrected state estimation and applying the homography to the commanded trajectories is simpler to implement and does not exhibit reduced performance.

Our approach can generalize to arbitrary transformations (not just homographies).  Triangular mesh warping, for example, may be more generalize-able to non-rectangular workspaces.  However, because our canvas is naturally rectangular and we wish to reduce the number of segments where possible, we choose to use homographies.

\subsection{Artwork Design and Trajectory Generation} \label{ssec:artwork_design}

The artwork designed for the mural in Fig. \ref{fig:headline} was designed by Al-Haddad in Rhino, an architectural CAD software.  It was generated parametrically to create sets of ellipses as a reference to deconstructing the texts of Galileo, translating them to drawings then code before reconstructing them on the face of the library\footnote{Additional details available online at \href{https://library.gatech.edu/AIR}{https://library.gatech.edu/AIR}}.

% To convert the artwork to a painting trajectory, we must create infill paths and compute timestamps.
To convert the artwork to a painting trajectory, we create infill paths in Rhino from the vector graphics with 15cm stepover then retime them using
% To create infill paths, toolpaths were generated in Rhino from the vector graphics with 15cm stepover.
% The toolpaths were then converted to gcode and retimed using
\cite{Chen24icra_qopp} to ensure feasible dynamics before applying the iLQG algorithm described in Sec. \ref{ssec:cdpr_control}.  In general, we find that converting artworks into toolpaths using off-the-shelf CAM software, retiming using \cite{Chen24icra_qopp}, then running the trajectory with robot-specific trajectory tracking controllers is a robust, efficient workflow for painting tasks without complex brush models.

\subsection{Operation}

The overall painting process is as follows.  Since, as described in Sec. \ref{ssec:cdpr_hardware_design}, our robot is narrower than our canvas and we paint in columnar sections, after each section we remove and reinstall the tetrahedral clamps supporting the pulleys in the next location.  We then perform the calibration procedure described in Sec. \ref{subsec:calibration} and prepare the brush and paint for the robot to paint.  Upon uploading the precomputed iLQR gains, the robot first navigates to its starting location using the ``manual mode'' dual-space PID controller before proceeding to follow the state machine in Fig. \ref{fig:overall_state_machine}.  Upon completion, the robot is place back into ``manual mode'' to return the platform to the operator.

\addtolength{\textheight}{-0.5cm}

%%%%%%%%%%%%%
\section{Experimental Results} \label{sec:results}
We evaluate our contributions of cable routing, brush stabilizer, and calibration algorithms separately in addition to evaluating the overall system through the completed mural.

\subsection{Cable Routing} \label{ssec:exp_cable_routing}
% Our motivation to use pulleys to increase mechanical advantage instead of decreasing the winch diameter is based on the premise that cable length estimation is less accurate the smaller the winch diameter.  
% As a result, applying a 2:1 pulley reduction doubles the winch diameter while keeping the same ratio of motor torque to cable tension which should lead to improved accuracy.
As discussed in Sec. \ref{ssec:cdpr_hardware_design}, we theoretically expect that cable length estimation error should not change with winch diameter for a fixed angle change $\theta$.  To experimentally validate this,
% To evaluate the premise of whether a smaller winch diameter leads to less accurate cable length estimation,
we conduct a small-scale experiment with a single winch and a single cable.  For various winch diameters, we measure the change in cable length for a fixed 25-rotation change in motor angle, repeated 10 times.  We find that the repeatability in millimeters is consistent with our theoretical analysis, showing no clear trend with the winch diameter.  The results are shown in Table \ref{tab:winch_diameter_vs_repeatability}.

% We hypothesize this to be the case because cable length estimation error is primarily due to variations in the effective winch radius caused by previous layers of cable winding on the winch.  Because these variations are in absolute lengths, the change in circumference $\delta c=\pi\delta d$.

\begin{table}
  \centering
  \caption{Repeatability of cable length estimation\\for various winch diameters}
  \label{tab:winch_diameter_vs_repeatability}
  \renewcommand*{\arraystretch}{1.2}
  \begin{tabular}{@{\extracolsep{4pt}}c|ccc@{}}
    \multirow{2}{*}{\shortstack{Nominal winch\\diameter (mm)}}
                  & \multicolumn{3}{c}{Cable length change for 25 rotations} \\[0.1em]
                  \cline{2-4}
                  & mean (m) & std (m) & std (\%)\\
    \hline
    11.5  &   1.0220  &   0.0025  &    0.244  \\
    14.5  &   1.2214  &   0.0033  &    0.273  \\
    18.0  &   1.4854  &   0.0014  &    0.091  \\
    20.0  &   1.6163  &   0.0024  &    0.147
  \end{tabular}\\[1em]
  The repeatability, as measured by the standard deviation of the cable length measurement, is comparable in millimeters, but worse as a percentage of cable length dispensed for smaller winch diameters.
\end{table}
% 25 rots

% 11.5mm:
% 1.024m
% 1.022m
% 1.021m
% 1.022m
% 1.025m
% 1.027m
% 1.021m
% 1.020m
% 1.019m
% 1.019m

% 14.5mm:
% % 3.628 - 2.393 < clearly an outlier
% 3.628 - 2.401
% 3.628 - 2.403
% 3.628 - 2.404
% 3.628 - 2.405
% 3.628 - 2.408
% 3.628 - 2.407
% 3.628 - 2.409
% 3.628 - 2.411
% 3.628 - 2.411

% 18mm:
% 3.628 - 2.142
% 3.628 - 2.141
% 3.628 - 2.143
% 3.628 - 2.143
% 3.628 - 2.142
% 3.628 - 2.143
% 3.628 - 2.141
% 3.628 - 2.143
% 3.628 - 2.142
% 3.628 - 2.146

% 20mm:
% 3.628 - 2.008
% 3.628 - 2.009
% 3.628 - 2.008
% 3.628 - 2.011
% 3.628 - 2.014
% 3.628 - 2.014
% 3.628 - 2.014
% 3.628 - 2.013
% 3.628 - 2.013
% 3.628 - 2.013

\subsection{Brush Stabilizer} \label{ssec:exp_brush_stabilizer}

% 500g mass

% With stabilizer:
% 0.9cm

% W/o:
% no weight: 36, brush: 28, 500g: 11.5
% deflection: brush alone: 8cm, brush with 500g: 24.5cm

We evaluate the brush stabilizer's ability to maintain the brush's orientation by measuring the deflection applying a 500g mass to the tip of the brush with and without the stabilizer.  We find that the stabilizer reduces the deflection significantly from 165mm to 9mm.  Although 500g is a pessimistic estimate for the maximum lateral force the brush will experience, we find that the stabilizer is effective at reducing deflection.

\subsection{Calibration} \label{ssec:exp_calibration}

\begin{figure}
  \centering
  \includegraphics[width=0.8\linewidth]{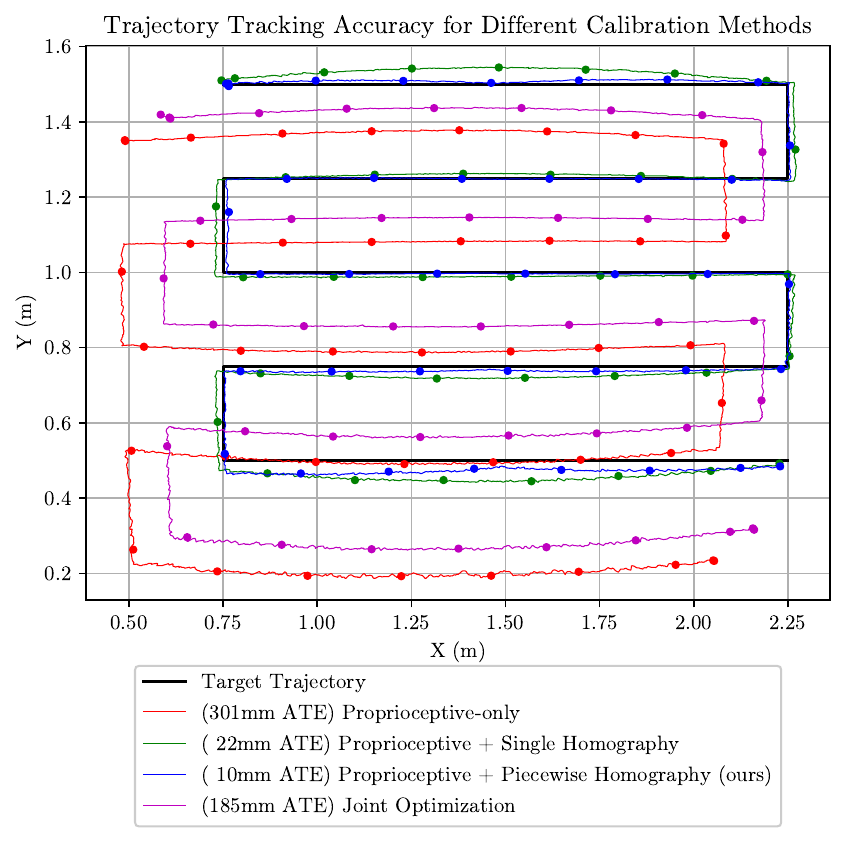}
  \caption{Tracking accuracy is significantly better when using our homography task-space calibration, even compared to including the same exteroceptive task-space calibration data in a joint optimization routine.}
  \label{fig:calibration_accuracy}
\end{figure}

We evaluate the 2-part calibration algorithm by measuring the trajectory tracking accuracy with ablations on a smaller scale (3m$\times$2.4m) testing CDPR.  Specifically, we evaluate the tracking accuracy after applying only the proprioceptive calibration, after applying one single homography to the entire workspace, and after applying a 2$\times$2 piecewise homography (4 sections, 9 grid points).  Additionally, we also evaluate an approach using a single, joint optimization problem for both proprioceptive and exteroceptive measurements\footnote{
  We formulate the optimization problem as a weighted nonlinear least squares problem with one term for the proprioceptive measurements as in \eqref{eq:calibration_optimization} and the other term using the exteroceptive measurements in place of $x_k$ in \eqref{eq:calibration_optimization}.  Various choices of weights were tried, but the results shown in this paper use weights of 94\% and 6\% for the proprioceptive and exteroceptive terms, respectively, normalized by number of elements.  Attempting higher weights for the exteroceptive term cause caused unsafe winch parameters.
} but find that it performs worse than our approach and only slightly better than the proprioceptive-only calibration.  Fig. \ref{fig:calibration_accuracy} shows the trajectories for each of the controllers as measured using AprilTag markers on the platform and robot frame, with the piecewise homography improving average tracking error (ATE) over proprioceptive-only from 301mm to 10mm.

\subsection{Mural} \label{ssec:exp_mural}

\begin{figure}
  \centering
  \scalebox{-1}[1]{\includegraphics[width=\linewidth]{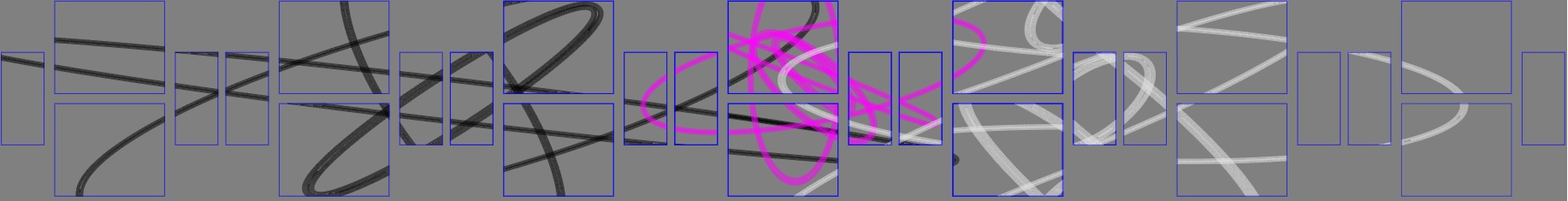}}
  % 235/1833*716 = 91.79487179 vs 152 -> 0.60391363
  % \vstretch{0.60391363}{
    \\[0.2em]
  \vstretch{0.90587045}{
    \includegraphics[width=0.99\linewidth]{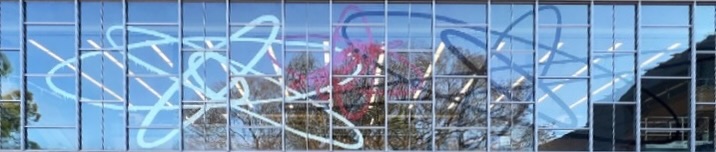}
  }
  \caption{Comparing the as-designed mural (top) to the completed, as-painted mural (bottom) demonstrates the proficiency of our robot.}
  \label{fig:simulated_vs_actual}
\end{figure}

We evaluate our overall robot system by painting the mural shown in Fig. \ref{fig:headline}.  We find that the robot is able to paint the mural with high accuracy and precision, and that the mural is well-received by the public.
Fig. \ref{fig:simulated_vs_actual} also compares the desired mural with the finished mural.

% We also compare a photo of the finished mural to a rendering of the mural to evaluate the quality of the robot's painting.  We find that the robot's painting is of high quality and closely resembles the rendering.
%\todo{do this.  Try to use IOU as a metric}

%%%%%%%%%%%%%
% \section{Discussion} \label{sec:discussion}
% \input{_discussion}

%%%%%%%%%%%%%
\section{Conclusions and Future Works} \label{sec:conclusions}
% \begin{figure}
%   \centering
%   \includegraphics[width=\linewidth,trim=0 150px 0 120px,clip]{figs/MuralOutside.png}
%   \caption{The complete mural, as viewed from the outside.}
%   \label{fig:mural_outside}
% \end{figure}

In this work, we created a robot system that can paint large-scale murals with a brush.
We combined the large workspace of a CDPR with the precision and dexterity of a serial manipulator to create a robot that can achieve the precision needed for touching a brush to the canvas across a large area.  We discussed a number of features that were required to attain sufficient accuracy, including novel ideas for cable routing, brush stabilization, and calibration, with experimental validation.
Finally, we evaluated the robot by publicly painting a large-scale 27m by 3.3m mural with positive reception from the public.
We believe that this work is a significant step towards more widespread and immersive robot art, and that it will inspire future work in this area.

%%%%%%%%%%%%%%%%%%%%%%%%%%%%%%%%%%%%%%%%%%%%%%%%%%%%%%%%%%%%%%%%%%%%%%%%%%%%%%%%

% \addtolength{\textheight}{-10.1cm}   % This command serves to balance the column lengths


\section*{ACKNOWLEDGMENT}
We graciously thank Catherine Manci for coordinating the Library's Artist-In-Residence program and Georgia Tech Mechanical Engineering undergraduate capstone teams %consisting of \ldots
for their contributions constructing the serial manipulator.

%%%%%%%%%%%%%%%%%%%%%%%%%%%%%%%%%%%%%%%%%%%%%%%%%%%%%%%%%%%%%%%%%%%%%%%%%%%%%%%%

\bibliographystyle{IEEEtran}
\bibliography{references/IEEEabrv,references/topp,references/common,references/mine,references/cdpr,references/PersonalLiterature,references/generative,references/chen22icra_painting,references/extra}
% \printbibliography

\end{document}